\documentclass[preprint,review,12pt]{elsarticle}

\usepackage{graphicx}

\usepackage{amssymb}

\usepackage{caption}

\usepackage[top=3.4cm,left=3.3cm,right=3.3cm,bottom=3.4cm]{geometry}

\usepackage{enumitem} 
  \setlist{itemsep=1ex plus0.2ex, leftmargin=*, align=left}

\makeatletter
\newcommand{\labitem}[2]{%
\def\@itemlabel{\textbf{#1}}
\item
\def\@currentlabel{#1}\label{#2}}
\makeatother

\makeatletter
\newcommand{\headingitem}[1]{%
\vspace{0.3cm}
\def\@itemlabel{\textbf{#1}}
\item
\def\@currentlabel{#1}
\addtocounter{enumi}{-1}
}
\makeatother

\usepackage{csvsimple}
\usepackage{multirow}
\usepackage{lipsum}

\usepackage{eurosym}
\usepackage{siunitx}

    \DeclareSIUnit\eur{\officialeuro}
    \DeclareSIUnit\M{M}
    \DeclareSIUnit\k{k}

  \def\sym#1{\ifmmode^{#1}\else\(^{#1}\)\fi}

\usepackage{etoolbox}
\robustify\bfseries

\usepackage{setspace} 
\usepackage{csquotes}
\usepackage{amsmath}
\usepackage{bm}

\usepackage{xspace}
	\newcommand\ie{i.\,e.\xspace}
	\newcommand\eg{e.\,g.\xspace}

	\newcommand\cf{cf.\xspace}

	\newcommand\US{U.\,S.\xspace}

\usepackage[amsmath,hyperref,framed]{ntheorem} 
  \theoremstyle{plain}
     
  \theoremstyle{nonumberplain}
    \theoremseparator{.}
    \theoremheaderfont{\bfseries}
    \theorembodyfont{\normalfont}
    \theoremsymbol{$\blacksquare$}
      \RequirePackage{amssymb}

      \makeatletter
    \let\copy@theorem@headerfont=\theorem@headerfont
    \newcommand{\my@theorem@headerfont}{%
        \boldmath\copy@theorem@headerfont\unboldmath
      }
    \let\theorem@headerfont=\my@theorem@headerfont
      \makeatother

\theoremstyle{nonumberplain}
\theoremseparator{.}
\newcommand{\argmax}{\operatornamewithlimits{arg \, max}}

\usepackage[%
  sort&compress
]{cleveref}
\usepackage{url}

\usepackage{algorithm}
\usepackage{algpseudocode}

\usepackage[usenames,dvipsnames]{xcolor}

\usepackage{colortbl}
\usepackage{tabularx}
\usepackage{booktabs}
\usepackage{collcell}

\usepackage{ragged2e}
  
\newcommand{\PreserveBackslash}[1]{\let\temp=\\#1\let\\=\temp}
\newcolumntype{v}[1]{>{\PreserveBackslash\RaggedRight\hspace{0pt}}p{#1}}

  \usepackage{adjustbox}
\newcolumntype{Q}[2]{%
    >{\adjustbox{angle=#1,lap=\width-(#2)}\bgroup}%
    l%
    <{\egroup}%
}

\newcommand{\mcellt}[2][c]{%
  \begin{tabular}[t]{@{}#1@{}}#2\end{tabular}}

\usepackage[
    colorinlistoftodos,
    textsize=footnotesize,
        ]{todonotes}

\makeatletter
    \renewcommand{\fps@figure}{htb}         
    \renewcommand{\fps@table}{htb}         
\makeatother 

\usepackage{float}
\usepackage{rotating}

\journal{Decision Support Systems}

\begin{document}

\begin{frontmatter}



\title{Decision support from financial disclosures\\ with deep neural networks and transfer learning}


\author[Freiburg]{Mathias Kraus\corref{cor1}}
\ead{mathias.kraus@is.uni-freiburg.de}

\author[Freiburg,ETH]{Stefan Feuerriegel}
\ead{sfeuerriegel@ethz.ch}

\address[Freiburg]{Chair for Information Systems Research, University of Freiburg, Platz der Alten Synagoge, 79098 Freiburg, Germany}
\address[ETH]{ETH Zurich, Weinbergstr. 56/58, 8092 Zurich, Switzerland}
\cortext[cor1]{Corresponding author. Mail: mathias.kraus@is.uni-freiburg.de; Tel: +49\,761\,203\,2395; Fax: +49\,761\,203\,2416.}

\begin{abstract}
Company disclosures greatly aid in the process of financial decision-making; therefore, they are consulted by financial investors and automated traders before exercising ownership in stocks. While humans are usually able to correctly interpret the content, the same is rarely true of computerized decision support systems, which struggle with the complexity and ambiguity of natural language. A possible remedy is represented by deep learning, which overcomes several shortcomings of traditional methods of text mining. For instance, recurrent neural networks, such as long short-term memories, employ hierarchical structures, together with a large number of hidden layers, to automatically extract features from ordered sequences of words and capture highly non-linear relationships such as context-dependent meanings. However, deep learning has only recently started to receive traction, possibly because its performance is largely untested. Hence, this paper studies the use of deep neural networks for financial decision support. We additionally experiment with transfer learning, in which we pre-train the network on a different corpus with a length of 139.1 million words. Our results reveal a higher directional accuracy as compared to traditional machine learning when predicting stock price movements in response to financial disclosures. Our work thereby helps to highlight the business value of deep learning and provides recommendations to practitioners and executives.
\end{abstract}


\end{frontmatter}



\section{Introduction}
\label{sec:Introduction}


The semi-strong form of the efficient market hypothesis states that asset prices adapt to new information entering the market~\cite{Fama.1965}. Included among these information signals are the regulatory disclosures of firms, as these financial materials trigger subsequent movements of stock prices \cite{Fisher.2016,Kearney.2014,Li.2010b,Loughran.2016}. Hence, investors must evaluate the content of financial disclosures and then decide upon the new valuation of stocks. Here a financial decision support system can greatly facilitate the decision-making of investors subsequent to the disclosure of financial statements \cite{Feuerriegel.2016b,deFortuny.2014,Geva.2014,Schumaker.2009,Schumaker.2009b,Schumaker.2012}. Corresponding decision support systems, such as those used by automated traders, can thereby help identify financially rewarding stocks and exercise ownership.

\begin{figure}[H]
\makebox[\textwidth]{%
\includegraphics[width=\textwidth]{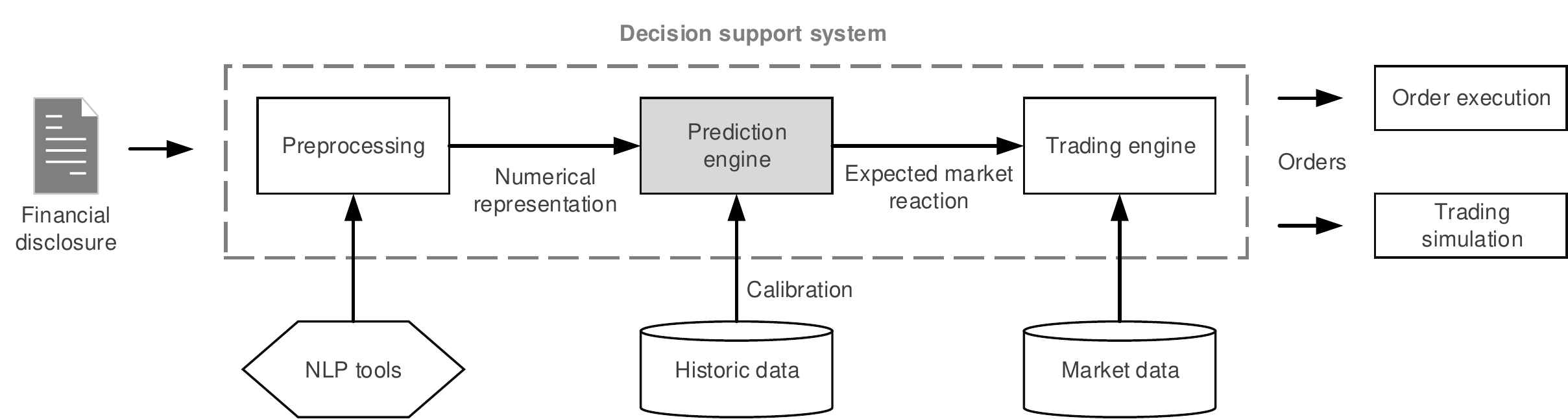}
}
\caption{Schematic illustration of a decision support system for news-based trading. The components are adapted from the systems described in \cite{Feuerriegel.2016,Geva.2014,Schumaker.2009,Schumaker.2009b,Schumaker.2012}. Here NLP refers to natural language processing. This work focuses on improvements to the underlying prediction engine.}
\label{fig:DSS}
\end{figure}

Decision support systems for news-based trading commonly consist of different components \cite{Feuerriegel.2016b,Geva.2014,Schumaker.2009,Schumaker.2009b,Schumaker.2012}, as schematically illustrated in \Cref{fig:DSS}. On the one hand, they need to assess the information encoded in the narratives of financial disclosures. For this purpose, a decision support system must rate the content of such disclosures in order to identify which stock prices are likely to surge or decrease. In other words, the system must quantify whether a financial disclosure conveys positive or negative content. For example, a prediction engine can forecast the expected price change subsequent to a disclosure. Afterwards, the trading engine decides whether to invest in a stock given the market environment. It also performs risk evaluations, and, if necessary, applies changes to the portfolio. The resulting financial performance of the portfolio largely depends upon the accuracy of the prediction engine, which constitutes the focus of this manuscript. Here even small improvements in prediction performance directly link to better decision-making and thus an increase in monetary profits.

Mathematically, the prediction takes a document $d \in D$ as input and then returns either the expected (excess) return or a class label denoting the direction of the price change (\ie positive or negative). Here a document $d$ is expressed as an ordered list $\left[ w_1, w_2, \ldots, w_m \right]$ of words, where the length $m$ of this list differs from document to document. When utilizing a traditional predictive model from machine learning (such as a support vector machine), the ordered list of length $m$ is mapped onto an vector of length $N$ that serves as input to the predictor. Independent of the varying length $m$, this predictor always entails the same length $N$. For this mapping operation, one primarily follows the bag-of-words approach~\cite{Manning.1999,Pang.2008}, which we detail in the following.

The bag-of-words approach~\cite{Manning.1999} counts the frequency of words (or tuples, so-called $n$-grams), while neglecting the order in which these words (or tuples) are arranged. Hence, this approach does not take into account whether one word (or $n$-gram) appears before or after another.\footnote{For instance, let us consider the examples \emph{\textquote{The decision was not good; it was actually quite bad}} and \emph{\textquote{The decision was not bad; it was actually quite good}}. When counting only the frequency of individual words, one cannot recognize that negation that changes the meaning of \emph{\textquote{good}} and \emph{\textquote{bad}}. Similar examples can also be constructed for $n$-grams.} Accordingly, the bag-of-words approach loses information concerning the meaning in a specific context~\cite{Nassirtoussi.2014,Pang.2008,Ravi.2015} and thus struggles with very long context dependencies that might span several sentences or paragraphs. The aforementioned properties underlying the bag-of-words approach are likely to explain why the accuracy of predictive models forecasting stock price movements on the basis of financial narratives is often regarded unsatisfactory~\cite{Nassirtoussi.2014}.

As an alternative to the bag-of-words approach, this paper utilizes recent advances in deep learning or, more precisely, sequence modeling based on deep neural networks. When applied to our case, these models allow us to consider very long context dependencies~\cite{Goodfellow.2017}, thereby improving predictive power. The underlying reason is as follows: deep neural networks for sequence modeling iterate over the running text word-by-word while learning a lower-dimensional representation of dimension $n$. By processing the text word-for-word, this approach preserves the word order and incorporates information concerning the context. Moreover, deep learning provides a very powerful framework when using large datasets in a variety of predictive tasks, as it is capable of modeling complex, non-linear relationships between variables or observations \cite{Goodfellow.2017}. Among the popular variants of deep neural networks for sequence modeling are recurrent neural networks~(RNNs) and the long short-term memory~(LSTM) model. Sequence modeling has successfully demonstrated the ability to store even long-ranging context information in the weights of network \cite{Goodfellow.2017}. For instance, in the related field of process mining, LSTMs have proven capable of effectively learning long sequences \cite{Evermann.2017}. Hence, sequence modeling also promises improvements to the predictive power of decision support systems for news-based trading. 

Despite recent breakthroughs in deep learning, our literature survey reveals that this approach is seldom employed in financial decision support systems. This gives one the impression that the business value of deep learning in practical applications is generally unknown. Accordingly, this paper sets out to address the following two research questions: (1)~Can sequence modeling with deep learning improve the prediction of short-term price movements subsequent to a financial disclosure as compared to bag-of-words approaches? (2)~Can we further bolster the predictive performance of deep neural networks by applying transfer learning?

As a primary contribution to the existing body of research, this work utilizes deep learning to predict stock returns subsequent to the disclosure of financial materials and evaluates the predictive power thereof. Each of the neural networks entails more than \num{500000} parameters that empower various variants of non-linearities. We then validate our results using an extensive collection of baseline methods, which are state of the art for bag-of-words. In addition, we tune the performance of our methods by applying concepts from transfer learning. That is, we perform representation learning (\ie pre-training word embeddings) using a different, but related, corpus with financial language and then transfer the resulting word embeddings to the dataset under study. Based on our findings, we provide managerial guidelines and recommendations for deep learning in financial decision support.

The remainder of this paper is structured as follows. \Cref{sec:related_work} provides an overview of related works on both decision support from financial news and deep learning for natural language processing. \Cref{sec:methods} then presents our baseline models, our network architectures, and our approach to transfer learning. These are utilized in \Cref{sec:results} to evaluate how deep learning can improve decision support in finance. Finally, \Cref{sec:discussion} discusses our findings and highlights the implications of our work for research and management. \Cref{sec:conclusion} concludes. 

\section{Related work}
\label{sec:related_work}

\subsection{Decision support from financial news}
\label{sec:financial_news}

Decision support from financial news is either of an explanatory or predictive nature. The former tries to explain the relationship between financial news and stock prices based on historic data. Specifically, this field of research utilizes econometrics in order to quantify the impact of news, establish a causal relationship between news and stock prices or understand which investors respond to news and how. Recent literature surveys~\cite{Fisher.2016,Kearney.2014,Li.2010b,Loughran.2016} provide a detailed overview of explanatory works, finding that these commonly count the instances of polarity cues, predefined by manually created dictionaries. 

In contrast, predictive approaches forecast the future reception of financial news by the stock market~\cite{Nassirtoussi.2014,Ravi.2015}. For this purpose, decision support systems are trained on historic data with the specific objective of performing accurately and reliably on unseen news. The resulting directional accuracy is usually only marginally better than \SI{50}{\percent} (and often merely on a subset of the original dataset), which demonstrates how challenging this task is. The approaches vary in terms of the underlying data source and predictive model, which we discuss in the following paragraphs. 

The text source can come in various forms such as, \eg, headlines of news stories~\cite{Huang.2010}, the content of newspaper articles~\cite{Wuthrich.1998,Schumaker.2012,Li.2014,Li.2014b,Chan.2016} or company-specific disclosures~\cite{Prollochs.2016,Feuerriegel.2016b}. In addition, research widely conducts numerical experiments with financial prices in daily resolution \cite{Nassirtoussi.2014}, and we adhere to this convention. For other resolutions, we refer to earlier works \cite[e.\,g.][]{Groth.2011} that further analyze the time dimension of news reception, including latency and peak effects in intraday trading.

In order to insert natural language into predictive models, the bag-of-words approach is widely utilized for the purpose of extracting numeric representations from the textual content \cite{Nassirtoussi.2014}. Commonly, additional scalings are applied, such as term frequency-inverse document frequency~(tf-idf). Accordingly, this study also utilizes the tf-idf approach for bag-of-words models, as our experiments demonstrate superior performance as compared to word frequencies. This numerical representation is then fed into predictive models, which are then trained on historic news and stock prices. Prevalent predictive models for financial news are na\"ive Bayes, regressions, support vector machines and decision trees~\cite{Nassirtoussi.2014}. All these methods perform well in situations with few observations and many potential regressors. Even though deep learning has achieved impressive results in natural language processing~\cite{Socher.2013,Pennington.2014}, this type of predictive model has been largely neglected in financial text mining. A noteworthy exception is the work in~\cite{Feuerriegel.2016}; however, it implements a very early form of deep learning -- namely, recursive autoencoders -- which can store context-sensitive information only for the course of a few words and is thus limited to learning simple semantics. The SemEval-2017 competition is currently raising awareness in this regard \cite{Cortis.2017}.

\subsection{Natural language processing with deep learning}
\label{sec:deep_learning_NLP}

For a long time, text mining was dominated by traditional machine learning methods, such as support vector machines, trained with high-dimensional yet very sparse feature vectors. It is only recently that researchers in the field of natural language processing have started to adapt ideas from advances in deep learning~\cite{Socher.2013,Pennington.2014}.\footnote{A comprehensive overview on deep learning for natural language processing is given in the tutorial of Richard Socher and Christopher Manning held at NAACL~HLT, 2013. URL: \url{http://nlp.stanford.edu/courses/NAACL2013/}, visited on July~6, 2017.} The utilized deep learning techniques are described in detail in~\cite{YoavGoldberg.2016,Goodfellow.2017}. 


The recurrent neural network processes raw text in sequential order~\cite{Williams.1986}. The connections in the neural network form a direct cycle, which allows for the passing of information from one word to the next. This helps the RNN to implicitly learn context-sensitive features. However, the RNN is subject to drawbacks (vanishing gradient problem and short context dependencies), which often prohibit its application to real-world problems~\cite{Bengio.1994}.

An improvement to the classical RNN is represented by the long short-term memory model, which is capable of processing sequential inputs with very long dependencies between related input signals~\cite{Hochreiter.1997}. For this purpose, the LSTM utilizes forget gates that prevent exploding gradients during back-propagation and thus numerical instabilities. As a consequence, LSTMs have become state of the art in many fields of research \cite{Goodfellow.2017} and we thus apply this deep learning architecture in our study on financial decision support.


\subsection{Deep learning in financial applications}
\label{sec:deep_learning_finance}

Despite being a very powerful framework, deep learning has rarely been used in finance research. Among the few such instances, financial time series prediction is one popular application. Here previous research utilizes an autoencoder composed of stacked restricted Boltzmann machines in order to predict future stock prices based on historic time series~\cite{Takeuchi.2013}. Similarly, the LSTM is applied to predict future stock returns and generates a portfolio of stocks that can yield higher returns than the top 10 stocks in the S\&P 500~\cite{Heaton.2016}.

Recent literature surveys~\cite{Nassirtoussi.2014,Ravi.2015} do not mention works that utilize deep learning for financial text mining; yet we found a two-stage approach that extracts specific word triples consisting of an actor, an action and an object from more than \num{400000} headlines of Bloomberg news and then applies deep learning in order to predict stock price movements \cite{Ding.2015}. However, this setup diminishes the advantage of deep learning, as it computes word tuples instead of processing the raw text. Closest to our research is the application of a recursive autoencoder to the headlines of approximately \num{6500} financial disclosures~\cite{Feuerriegel.2016}. This early work trains a recursive autoencoder and uses the final code vector to predict stock price movements. However, this network architecture can rarely learn long context dependencies \cite{Goodfellow.2017} and thus struggles with complex semantic relationships. Furthermore, the dataset is subject to extensive filtering in order to yield a performance that, ultimately, is only marginally better than random guessing. 

Based on our review, we are not aware of any works that utilize recent advances in deep learning -- namely, recurrent neural networks or LSTMs -- in order to improve decision support based on financial news. 

\section{Methods and materials}
\label{sec:methods}

\begin{figure}[H]
\makebox[\textwidth]{%
\includegraphics[width=0.85\textwidth]{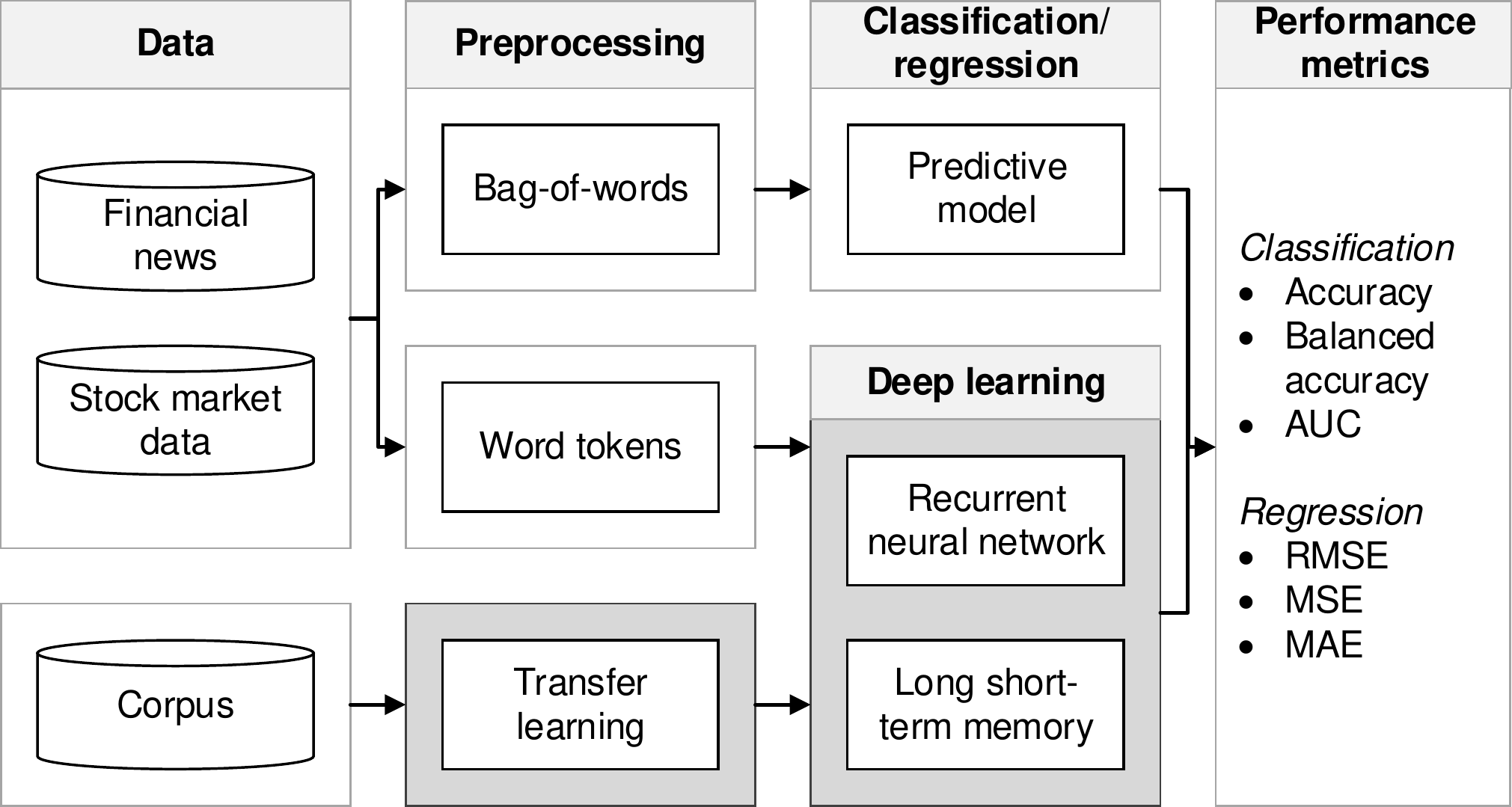}
}
\caption{Research framework evaluating the performance gains from deep learning architectures and transfer learning.}
\label{fig:framework}
\end{figure}

This section introduces our methodology, as well as the dataset, to predict stock price movements on the basis of financial disclosures. In brief, we compare na\"ive machine learning using bag-of-words with novel deep learning techniques (see \Cref{fig:framework}).

We specifically experiment with (a)~classification, where we assign the direction of the stock price movement -- \emph{up} or \emph{down} -- to a financial disclosure, and (b)~regression, where we predict the magnitude of the change. In both cases, we study price changes in terms of both nominal returns and abnormal returns. The latter corrects returns for confounding market movements and isolates the effect of the news release itself.  

\subsection{Dataset}
\label{sec:datasets}


Our corpus comprises \num{13135} regulated German ad~hoc announcements in English.\footnote{These disclosures are publicly available via the website of the \textquote{DGAP} (\url{www.dgap.de}). Moreover, the specific dataset of this study can be downloaded from \url{https://github.com/MathiasKraus/FinancialDeepLearning}.} This type of financial disclosure is an important source of information, since listed companies are obliged by law to publish these disclosures in order to inform investors about relevant company occurrences (\cf German Securities Prospectus Act). They have shown a strong influence on financial markets and have been a popular choice in previous research~\cite{Prollochs.2016,Groth.2011}.

Consistent with previous research, we reduce noise by omitting the disclosures of penny stock companies, \ie those with a stock price below \EUR{5}. In addition, we only select disclosures published on trading days. This yields a sample of \num{10895} observations. We compute abnormal returns with daily stock market data using a market model whereby the market is modeled via the CDAX during the 20 trading days prior to the disclosure. In the classification task, we label each disclosure as \emph{positive} (encoded as \num{+1}) or \emph{negative} (encoded as \num{0}) based on the sign of the corresponding return. 


\Cref{tbl:summary_statistics} shows descriptive statistics. Ad~hoc releases are composed of \num{168.80} words on average, which is significantly longer than the documents used in most applications of deep learning in the field of natural language processing. In total, \num{12444} unique words appear in our dataset. The distributions of both abnormal and nominal return are substantially right-skewed, as indicated by the \SI{25}{\percent} and \SI{75}{\percent} percentiles. Our dataset contains a few market movements of larger magnitude due to a number of factors, including mergers and acquisitions, as well as bankruptcy. We deliberately include these values in our data sample as a decision support system in live application can neither recognize nor filter them. We further observe an unbalanced set of labels, a fact which needs to be adjusted for in the performance measurements. For the nominal return, a positive label appears \SI{9}{\percent} more frequently than a negative label.

\begin{table}
\footnotesize
\makebox[\textwidth]{%
    \begin{tabular}{l S[table-format=5]SSSSSSSS}
		    \toprule
    	  {\textbf{Variable}} &
        {\textbf{Obs.}} & 
        {\textbf{Mean}} &        
	    	{\textbf{Std. dev.}} & 
        {\textbf{Coef. of}} & 
				\multicolumn{5}{c}{\textbf{Percentiles}} \\
				\cmidrule(l){6-10}
				& & & & {\textbf{var.}} & 
        {\textbf{5\,\%}} & 
        {\textbf{25\,\%}} & 
        {\textbf{50\,\%}} & 
        {\textbf{75\,\%}} &
        {\textbf{95\,\%}} \\ 
        \midrule
        Abnormal return & 10895 & 0.699 & 6.923 & 9.908 &
         -7.639 & -1.712 & 0.223 & 2.749 & 10.010 \\
        Nominal return & 10895 & 0.778 & 6.876 & 8.835 &
         -7.410 & 1.530 & 0.220 & 2.757 & 9.961 \\
	      Length (in words) & 10895 & 168.801 & 117.397 &
		0.695 & 61.000 & 95.000 & 137.000 & 203.000 &
		 381.000 \\
        \bottomrule
    \end{tabular}
}
\caption{Summary statistics of the stock market data, as well as the length of disclosures.}
\label{tbl:summary_statistics}
\end{table}


To measure the predictive performance, we split the dataset into a training and a test set. The first \SI{80}{\percent} of the time frame gives our training data, while the last \SI{20}{\percent} defines our test set. This differs from earlier studies which appear to draw hold-out samples from the same time period as the training set~\cite{Lau.1987}. As a drawback, the latter process ignores the chronological order of disclosures and, hence, training would erroneously benefit from data samples that otherwise are only ex post available.\footnote{As an example, let us consider a case where the training/test data is not split in chronological order. For instance, data from 2010 (\ie after the financial crisis) is used for training, while data from 2008/09 (\ie during the financial crisis) is used for testing. The algorithm would then learn that the term \textquote{Lehman Brothers} has a negative connotation and, as a result, it would accurately predict the bankruptcy of Lehman Brothers, since it had ex post knowledge that a decision support system would not have had in a real-world setting.} We thus follow \cite{Faraway.2016,SerranoCinca.2015} and split the training and test sets in chronological order. A similar approach is applied in cross-validation, as detailed later. After splitting, the dataset contains \num{8716} disclosures for training and \num{2179} for testing.

\subsection{Baselines with bag-of-words}
\label{sec:baselines}

This section briefly describes baseline models based on bag-of-words. First, we tokenize each document, convert all characters to lower case and remove punctuation as well as numbers. Moreover, we perform stemming~\cite{Manning.1999}, which maps inflected words onto a base form; \eg \emph{\textquote{increased}} and \emph{\textquote{increasing}} are both mapped onto \emph{\textquote{increas}}~\cite{Porter.1980}. Since we a priori cannot know whether stemming actually improves the predictive performance, we later incorporate this decision as a tuning parameter. Thereby, stemming is only utilized when it actually improves the predictive performance on the validation set. We then transform the preprocessed content into numerical feature vectors by utilizing the tf-idf approach, which puts stronger weights on characteristic terms~\cite{Manning.1999}. Furthermore, we report the results from using unigrams as part of our evaluation. In addition, we later perform a sensitivity analysis and incorporate short context dependencies by employing sequences of adjacent words up to length $n$ to form $n$-grams. 



The selection of baseline predictors includes linear models, such as ridge regression, lasso and elastic nets, as well as non-linear models, such as random forest, AdaBoost and gradient boosting. We also employ support vector machines with both linear and non-linear kernels~\cite{Wang.2012}. These models have been shown to perform well on machine learning problems with many features and few observations~\cite{Hastie.2009}; hence, they are especially suited to our task, where the number of predictors exceeds the number of documents. 

\subsection{Deep learning architectures}
\label{sec:DL}

We first introduce the RNN, followed by its extension, the LSTM, which can better memorize information \cite{Goodfellow.2017}. Both network architectures iterate over sequential data $x_1, x_2, \ldots$ of arbitrary length. Here the input vector $x_i$ consists of the words (or stems) in one-hot encoding. Mathematically this specifies a vector consisting of zeros, except for a single element with a 1 that refers to the $i$-th word in the sequence \cite{Goodfellow.2017}. This yields high-dimensional but sparse vectors as input. In addition, we experiment with word embeddings in the case of the LSTM, as detailed below.


\subsubsection{Recurrent neural networks}
\label{sec:RNN}

\begin{figure}
\centering
\includegraphics[width=0.3\textwidth]{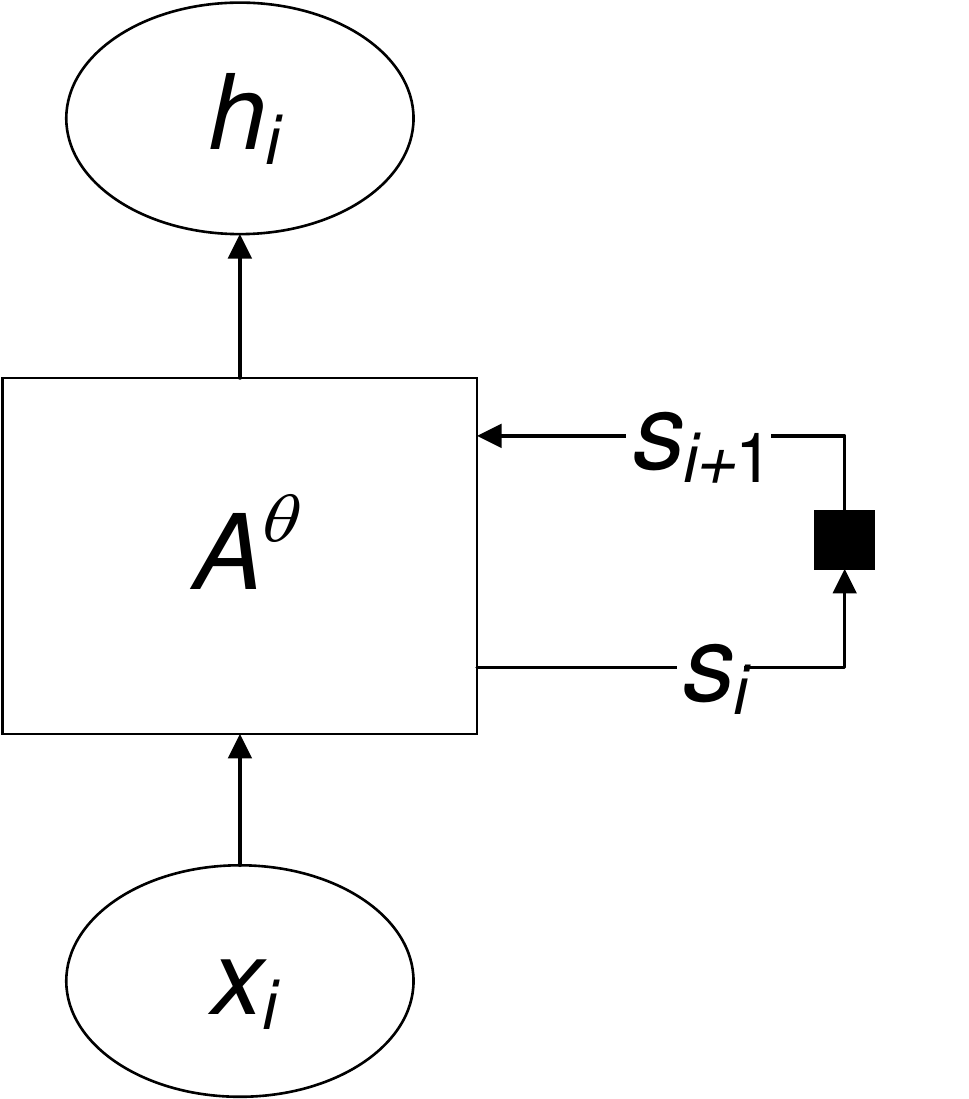}
\caption{Schematic structure of a recurrent neural network with input $x_i$, state $s_i$, output $h_i$ and one feedforward neural network $A^\theta$ parameterized by $\theta$. When moving from word $i$ to $i+1$, the recurrent neural network can pass information related to the current and previous words on by sending information from state $s_i$ to the next state $s_{i+1}$. It thereby draws upon previous terms and encodes context dependencies between words.}
\label{fig:RNN}
\end{figure}

The recurrent neural network~\cite{Williams.1986} allows the connections between neurons to form cycles, based on which the network can memorize information that persists when moving from word $x_i$ to $x_{i+1}$. The architecture of an RNN is illustrated in \Cref{fig:RNN}. 

\begin{figure}
\centering
\includegraphics[width=.7\textwidth]{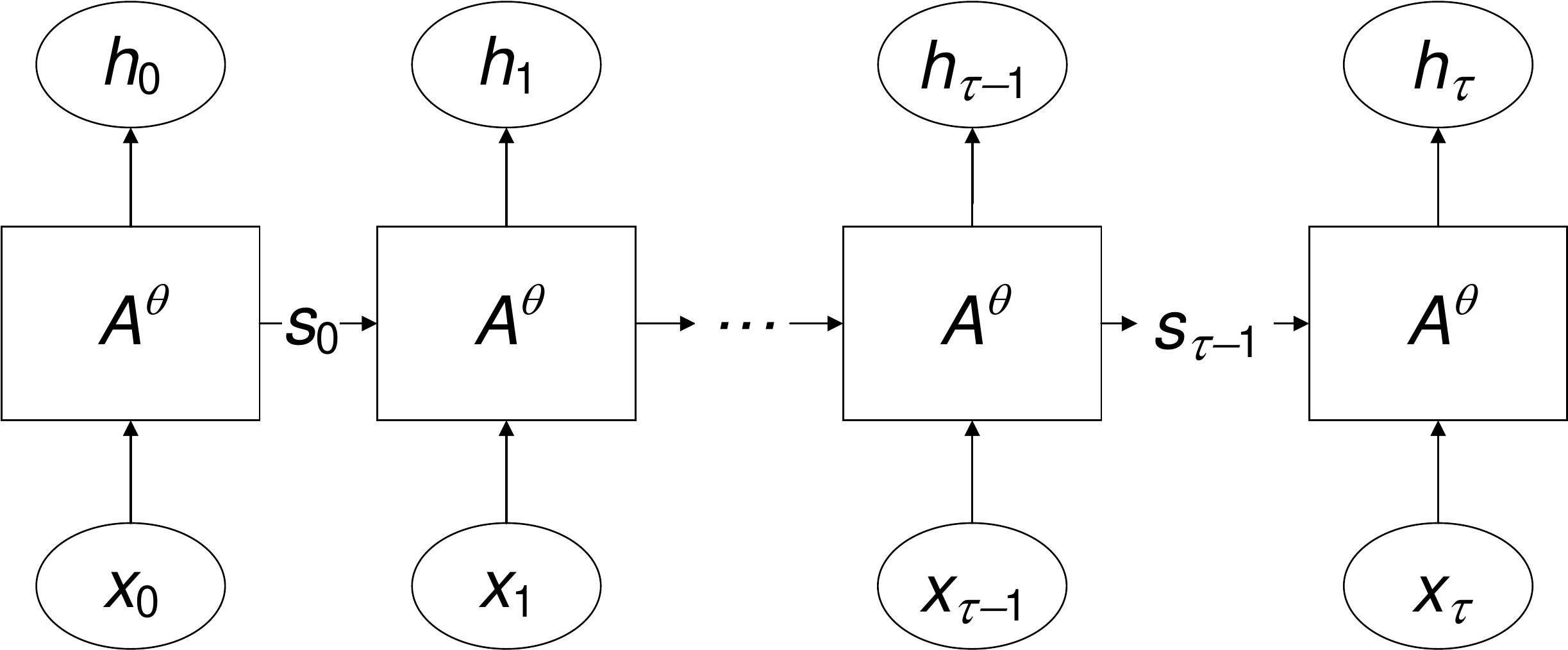}
\caption{Recurrent neural network unrolled over inputs $x_0, \ldots, x_\tau$, states $s_0, \ldots, s_{\tau-1}$, outputs $h_0, \ldots, h_\tau$ and a feedforward neural network $A^\theta$.}
\label{fig:RNN_unrolled}
\end{figure}

Let $x_i$ be the input in iteration $i$. Furthermore, $A^\theta$ denotes the feedforward neural network parameterized by $\theta$, while $s_i$ is the hidden state and $h_i$ is the output in iteration $i$. When moving from iteration $i$ to $i+1$, the RNN calculates the output $h_{i+1}$ from the neural network $A^\theta$, the previous state $s_i$ and the current input $x_{i+1}$, \ie
\begin{equation}
h_{i+1} = A^\theta(s_i, x_{i+1}).
\end{equation}
By modeling a recurrent relationship between states, the RNN can pass information onwards from the current state $s_i$ to the next $s_{i+1}$. To illustrate this, \Cref{fig:RNN_unrolled} presents the processing of sequential data by unrolling the recurrent structure. 

Theoretically, RNNs are very powerful, and yet two issues limit their practical application \cite{Goodfellow.2017}. First, vanishing and exploding gradients during training result in numerical instabilities and, second, information usually only persists for a few iterations in the memory~\cite{Bengio.1994}. 

\subsubsection{Long short-term memory networks}
\label{sec:LSTM}

\begin{figure}[H]
\centering
\includegraphics[width=1.0\textwidth]{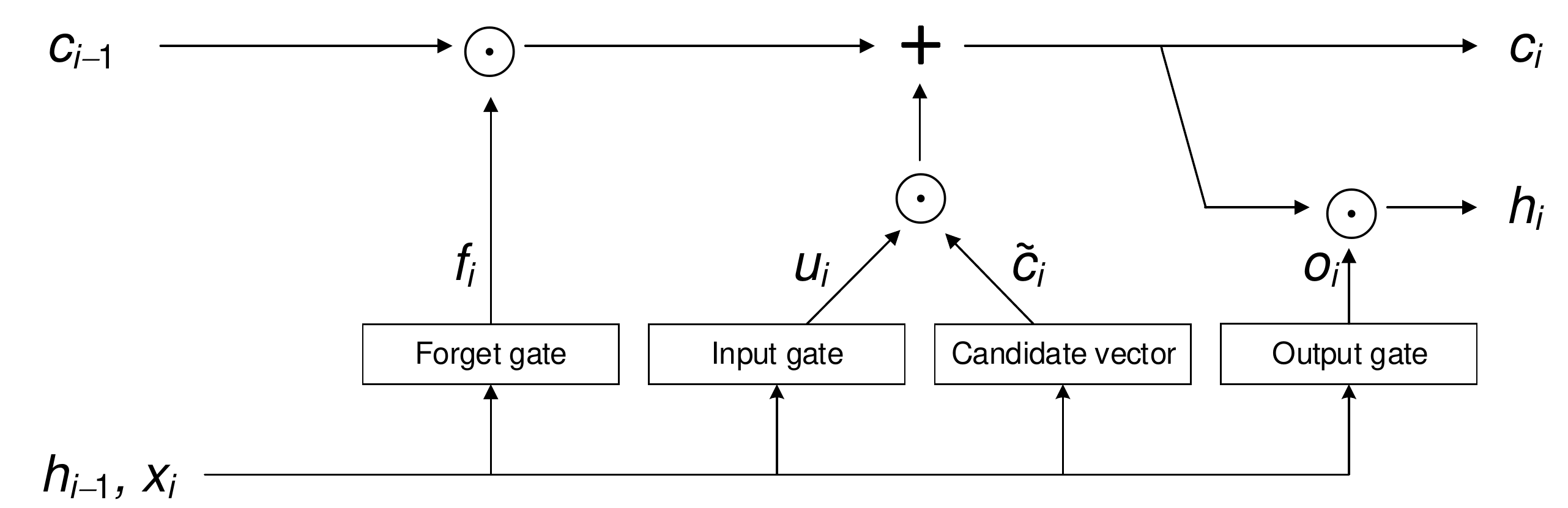}
\caption{Long short-term memory with input $x_i$, output $h_i$, cell state $c_i$ and four gates to filter information.}
\label{fig:LSTM}
\end{figure}

Long short-term memory networks advance RNNs to capture very long dependencies among input signals \cite{Goodfellow.2017}. For this purpose, LSTMs still process information sequentially, but introduce a cell state $c_i$, which remembers and forgets information, similar to a memory \cite{Hochreiter.1997}. This cell state is passed onwards, similar to the state of an RNN. However, the information in the cell state is manipulated by additional structures called gates. The LSTM has three of them -- namely the forget gate, the input gate and the output gate -- each of which is a neural network layer with its own sigmoid activation function. This is schematically visualized in \Cref{fig:LSTM}.

The forget gate takes the output $h_{i-1}$ from the previous word and the numerical representation $x_i$ of the current word as input. It then returns a vector $f_i$ with elements in the range $[0, 1]$. The values correspond to the strength with which each element in cell state $c_i$ should be passed on to the next cell state. Here a zero refers to discarding, a one to remembering. 

Next, we compute what information finds its way into the cell state. On the one hand, an input gate takes $h_{i-1}$ and $x_i$ as input and returns a vector $u_i$ denoting which elements in $c_{i-1}$ are updated. On the other hand, an additional neural network layer computes a vector of candidate values $\tilde{c}_i$ that might find its way into the cell state. Both are combined by element-wise multiplication, as indicated by the operator $\odot$.

Lastly, we need to define how a new cell state $c_i$ translates into an output $h_i$. This is accomplished with an output gate that computes a numeric vector $o_i$ with elements in the range $[0, 1]$. These values refer to the elements in $c_i$ which are passed on to the output. The new output is obtained through element-wise multiplication, \ie $h_i = o_i \odot c_i$. The new cell state stems from the updating rule
\begin{equation}
c_{i} = f_i \odot c_{i-1} + u_i \odot \tilde{c}_i.
\end{equation} 

In order to make predictions, the LSTM utilizes the final output $h_\tau$ and inserts this into an additional feedforward layer \cite{Goodfellow.2017}. Therefore, one simultaneously optimizes both the activation functions of the gates and this last feedforward layer with a combined target function. During training, we tune several parameters inside the LSTM (see \Cref{sec:tuning_parameters}). 

In practice, we not only insert binary vectors with words as one-hot encoding into the LSTM, but also utilize word embeddings~\cite{Pennington.2014,Mikolov.2013}. Word embeddings construct a lower-dimensional, dense vector representation of word tokens from the originally sparse and high-dimensional vectors, while preserving contextual similarity. We generate word embeddings by the GloVe algorithm~\cite{Pennington.2014} and, subsequently, fine-tune them together with the weights of the neurons during the actual training phase. 

\subsection{Model tuning}
\label{sec:tuning_parameters}

\Cref{alg:tuning} describes the tuning in order to find the best-performing parameters based on time-series cross-validation that employs a rolling forecasting origin~\cite{Hyndman.2014}. We first split the training data $\mathcal{T}$ into $k$ disjoint subsets $\mathcal{T}_1, \ldots, \mathcal{T}_k$ that are chronologically ordered. Afterwards, we set the tuning range for each parameter, iterate over all possible combinations and over all subsets of $\mathcal{T}$. In each iteration $i$, we measure the performance $\mathit{perf}_i$ of the method on the validation set $\mathcal{T}_i$ while using the subsets $\mathcal{T}_1, \ldots, \mathcal{T}_{i-1}$ from the previous points in time as training. Finally, we return the best-performing parameter setting. 

\begin{algorithm}[H]
\caption{Parameter tuning with time-series cross-validation}
\label{alg:tuning}
\footnotesize
\begin{algorithmic}
\State $\mathcal{T}\gets$ Training data which is chronologically ordered
\State Split $\mathcal{T}$ into $k$ disjoint subsets $\mathcal{T}_1, \ldots, \mathcal{T}_k$ by maintaining the chronological order
\State $\mathcal{R}_{i}\gets$ Ranges of tuning parameter $i$ with $i = 1, \ldots, l$
\For{$(p_1, \ldots, p_l)$ \textbf{in} $\mathcal{R}_1 \times \ldots \times \mathcal{R}_l$}
  \For{$i$ \textbf{in} $2,\ldots,k$}
    \State Train model $m$ with data from subsets $\mathcal{T}_1, \ldots, \mathcal{T}_{i-1}$
    \State $\mathit{perf}_i\gets$ Measure performance of model $m$ on $\mathcal{T}_i$
  \EndFor
  \State $\mathit{perf}_{(p_1, \ldots, p_l)} \gets \frac{1}{k}\sum\limits_{i = 1}^k \mathit{perf}_i$
\EndFor 
\State \textbf{return} $\argmax\limits_{(p_1, \ldots, p_l)}{\;\mathit{perf}_{(p_1, \ldots, p_l)}}$
\end{algorithmic}
\end{algorithm}

The tuning parameters are detailed in the online appendix. To tune all parameters of the baseline methods, we perform a grid search with \num{10}-fold time-series cross-validation on the training set. In the case of deep learning, we tune the parameters of architectures by using the last \SI{10}{\percent} of the training set for validation due to high computational demand. We optimize the deep neural networks by utilizing the Adam algorithm with learning rates tuned on the interval $[0.0001, 0.01]$ with a step size of 0.0005, while weights are initialized by the Xavier algorithm. The size of the word embeddings is tuned on the set $\{30, 40, \ldots, 100\}$. We initialize the word embeddings based on a continuous uniform distribution $\mathcal{U}(-0.1, 0.1)$ and set the size of each neural network layer within the RNN and LSTM to the dimension of the word embeddings.

\subsection{Transfer learning}
\label{sec:transfer_learning}

Transfer learning performs representation learning on a different, but related, dataset and then applies the gained knowledge to the actual training phase~\cite{Pan.2010}. In the case without transfer learning, the weights in the neural network are initialized randomly and then optimized for the training set. Given the sheer number of weights in deep neural networks, this approach requires a large number of training samples in order for the weights to converge. The idea behind transfer learning is to initialize the weights not randomly, but rather with values that might be close to the optimized ones. For this purpose, we utilize an additional dataset with 8-K filings and train the neural network (including word embeddings, if applicable) to predict stock price movements from this dataset. The resulting weights then serve as initial values when performing the actual training process for optimizing the weights with the ad~hoc announcements.

More explicitly, we draw upon \num{34782} Form~\num{8}-K filings, spanning the years 2010 to 2013, with a total length of \num{139.1}~million words.\footnote{These disclosures are publicly available via the website of the \US Securities and Exchange Commission (\url{www.sec.gov/edgar.shtml}).} This type of disclosure is mandated by the \US Securities and Exchange Commission to inform investors about stock-relevant events. Form~\num{8}-K filings contain considerably more words than ad~hoc announcements: they comprise an average of \num{4000.15} words, compared to a mean of \num{168.80} words in the case of ad~hoc announcements. The vocabulary of the \num{8}-K filings includes \num{57732} unique terms, out of which \num{7239} entries also appear in ad~hoc releases. Word embeddings for all terms not part of 8-K filings are drawn from a uniform distribution $\mathcal{U}(-0.1,\,0.1)\,$. 

\section{Results}
\label{sec:results}

This section compares the performance of bag-of-words and deep learning architectures on the basis of financial disclosures. The evaluation provides evidence that deep learning is superior to traditional bag-of-words approaches in predicting the direction and magnitude of stock price movements. In addition, our results clearly demonstrate additional benefits from using transfer learning in order to further bolster the performance of deep neural networks. 

Undertaking transfer learning is often computationally intensive, especially in cases with an extensive corpus such as ours. We thus utilized a computing system consisting of an Intel Xeon E5-2673 V3 with 8 cores running at \num{3.2}~GHz and \num{16}~GB~RAM. The training of LSTMs ranges below 5~hours, while the overall runtime, including transfer learning, amounted to approximately 22~hours.

We implemented all baselines in Python using scikit-learn, while we used TensorFlow and Theano for all experiments with deep learning. The resulting neural networks from the deep learning process are available for download via \url{https://github.com/MathiasKraus/FinancialDeepLearning}. This is intended to facilitate future comparisons and direct implementations in practical settings. All networks are shipped in the HDF5 file format as used by the Keras library.

We report the following metrics for comparing the performance of both regression and classification tasks. In case of the former, we compute the root mean squared error~(RMSE), the mean squared error~(MSE) and the mean absolute error~(MAE) in order to measure the deviation from the true return. Our classifications specifically compare the balanced accuracy, which is defined as the arithmetic mean of sensitivity and specificity, in order to account for unbalanced classes in our dataset. For the same reason, we also provide the area under the curve~(AUC). 

\subsection{Classification: direction of nominal returns}

We now proceed to evaluate the classifiers for predicting the direction of nominal returns. The corresponding results are detailed in~\Cref{tbl:classification_nominal_return}. The first row reflects the performance of our na{\"i}ve benchmark when using no predictor (\ie voting the majority class). It results in an accuracy above average due to the severe class imbalance. This also explains why we compare merely the balanced accuracy in the following analysis. Among the baseline models from traditional machine learning, we find the highest balanced accuracy on the test set when using the random forest, which yields an improvement of \num{4.7} percentage points compared to the na{\"i}ve benchmark. Its results stem from a random forest with \num{500} trees, where 3 variables are sampled at each split. This highlights once again the strength of the random forest as an out-of-the-box classifier. 

Deep learning outperforms traditional machine learning. For instance, the LSTM with word embeddings yields an improvement of \num{6.8} percentage points over the na{\"i}ve baseline. The word embeddings contribute to this increase by a mere \num{0.1} percentage points; however, they elevate the AUC score by \num{0.5} percentage points. 

Transfer learning yields consistent improvements for deep learning variants. As a result, the LSTM model with word embeddings performs best among all approaches, amounting to a total improvement of \num{7.1} percentage points. In other words, transfer learning enhances the balanced accuracy by an additional \num{0.8} percentage points. Compared to the strongest traditional model (AUC of \num{0.556}), transfer learning increases the AUC score by \num{0.021} (significant at the \num{0.05} level), thereby reaching an AUC of \num{0.577}. 

\begin{table}
\hspace{-2.8cm}
\footnotesize
    \begin{tabular}{l SSSS SSSS} 
		    \toprule
		{\textbf{Method}} & 
		{\textbf{Training set}} & 
		\multicolumn{3}{c}{\textbf{Test set}} &
		\multicolumn{3}{c}{\textbf{\mcellt{Absolute improvement\\ on test set over baseline}}} \\
 		\cmidrule(l){2-2} \cmidrule(l){3-5} \cmidrule(l){6-8}
		& {\textbf{Accuracy}} &
		{\textbf{Accuracy}} & 
		{\textbf{\mcellt{Balanced\\ accuracy}}} &
		{\textbf{AUC}} &
		{\textbf{Accuracy}} & 
		{\textbf{\mcellt{Balanced\\ accuracy}}} &
		{\textbf{AUC}} \\
        \midrule
				\multicolumn{7}{l}{\textsc{Na{\"i}ve baseline}} \\
        \quad Majority class & 0.549 & 0.540 & 
                             0.500 & 0.500 &
                             \text{--} & 
                             \text{--} & \text{--} \\   
        \midrule
        \multicolumn{7}{l}{\textsc{Traditional machine learning}} \\
        \quad Ridge regression & 0.534 & 0.534 & 
                                 0.528 & 0.539 &
                                 -0.006 & 
                                 0.028 & 0.039 \\
        \quad Lasso & 0.549 & 0.540 & 
                      0.500 & 0.500 & 
                      0.000 & 
                      0.000 & 0.000 \\ 
        \quad Elastic net & 0.549 & 0.540 & 
                            0.500 & 0.500 & 
                            0.000 & 
                            0.000 & 0.000 \\        
        \quad Random forest & 0.557 & 0.562 & 
                              0.547 & 0.552 & 
                              0.022 & 
                              0.047 & 0.052 \\
        \quad SVM  & 0.552 & 0.545 & 
                     0.522 & 0.556 &
                     0.005 & 
                     0.022 & 0.056 \\
        \quad AdaBoost & 0.555 & 0.552 & 
                         0.538 & 0.555 & 
                         0.012 & 
                         0.038 & 0.055 \\
        \quad Gradient boosting & 0.553 & 0.554 & 
                                  0.532 & 0.556 &
                                  0.014 & 
                                  0.032 & 0.056 \\
        \midrule
				\multicolumn{7}{l}{\textsc{Deep learning}} \\
        \quad RNN & 0.588 & 0.545 & 
                    0.530 & 0.529 &
                    0.005 & 
                    0.030 & 0.029 \\
        \quad LSTM & \bfseries 0.601 & \bfseries 0.577 & 
                     \bfseries 0.562 & \bfseries 0.563 &
                     \bfseries 0.037 & 
                     \bfseries 0.062 & \bfseries 0.063 \\
        \quad LSTM with word embeddings & \bfseries 0.597 & \bfseries 0.576 & 
                                          \bfseries 0.563 & \bfseries 0.568 & 
                                          \bfseries 0.036 & 
                                          \bfseries 0.063 & \bfseries 0.068 \\
				\midrule
				\multicolumn{7}{l}{\textsc{Transfer learning}} \\
        \quad RNN with pre-training & \bfseries 0.596 & 0.548 & 
                                      0.533 & 0.530 & 
                                      0.008 & 
                                      0.033 & 0.033 \\
				\quad LSTM with pre-training & \bfseries 0.576 & \bfseries 0.578 & 
                                       \bfseries 0.564 & \bfseries 0.577 & 
				                               \bfseries 0.038 & 
                                       \bfseries 0.064 & \bfseries 0.077 \\
				\quad LSTM with pre-training & \bfseries 0.581 & \bfseries 0.580 & 
                                       \bfseries 0.571 & \bfseries 0.568 &
				                               \bfseries 0.040 & 
                                       \bfseries 0.071 & \bfseries 0.068 \\
				\quad and word embeddings \\
        \bottomrule
    \end{tabular}
\caption{Out-of-sample results from classifying the direction of the nominal return. Values in bold indicate approaches that outperform the na{\"i}ve baseline and all models from traditional machine learning.}
\label{tbl:classification_nominal_return}
\end{table}

\subsection{Classification: direction of abnormal returns}

\Cref{tbl:classification_abnormal_return} reports the results for predicting the direction of abnormal returns, depicting a picture similar to that of the classification of nominal returns. The random forest again scores well with a balanced accuracy of \num{0.542}, but is beaten by \num{0.545} from the ridge regression. The latter achieves a balanced accuracy that is \num{4.5} percentage points higher than the na{\"i}ve benchmark. This performance is obtained when $\alpha$ is set to \num{0.99}.

With regard to deep learning, both RNNs (with and without transfer learning) fail to improve performance beyond traditional machine learning models. However, the LSTM still succeeds in this task, exceeding the balanced accuracy of the na{\"i}ve benchmark by \num{5.6} percentage points. Further performance gains come from the use of word embeddings, representing an improvement of \num{6.6} percentage points compared to the na{\"i}ve approach. 

When applying transfer learning, LSTMs show further, considerable improvements, since they outperform the balanced accuracies of the LSTMs without transfer learning by \num{0.7} and \num{1.7} percentage points, respectively. The LSTM with both pre-training and word embeddings further enhances this value to \num{8.3} percentage points.

\begin{table}
\hspace{-2.8cm}
\footnotesize
    \begin{tabular}{l SSSS SSSS} 
		    \toprule
		{\textbf{Method}} & 
		{\textbf{Training set}} & 
		\multicolumn{3}{c}{\textbf{Test set}} &
		\multicolumn{3}{c}{\textbf{\mcellt{Absolute improvement\\ on test set over baseline}}} \\
 		\cmidrule(l){2-2} \cmidrule(l){3-5} \cmidrule(l){6-8}
		& {\textbf{Accuracy}} &
		{\textbf{Accuracy}} & 
		{\textbf{\mcellt{Balanced\\ accuracy}}} &
		{\textbf{AUC}} &
		{\textbf{Accuracy}} & 
		{\textbf{\mcellt{Balanced\\ accuracy}}} &
		{\textbf{AUC}} \\
        \midrule
				\multicolumn{7}{l}{\textsc{Na{\"i}ve baseline}} \\
        \quad Majority class  & 0.542 & 0.528 & 
                            0.500 & 0.500 &
                            \text{--} & 
                            \text{--} & \text{--} \\   
        \midrule 
        \multicolumn{7}{l}{\textsc{Traditional machine learning}} \\
        \quad Ridge regression & 0.539 & 0.549 & 
                                 0.545 & 0.562 &
                                 0.021 & 
                                 0.045 & 0.062\\
        \quad Lasso & 0.542 & 0.528 & 
                      0.500 & 0.500 &
                      0.000 & 
                      0.000 & 0.000 \\ 
        \quad Elastic net & 0.542 & 0.528 & 
                            0.500 & 0.500 & 
                            0.000 & 
                            0.000 & 0.000 \\  
        \quad Random forest & 0.559 & 0.552 & 
                              0.542 & 0.559 &
                              0.024 & 
                              0.042 & 0.059 \\
        \quad SVM & 0.536 & 0.557 & 
                    0.527 & 0.558 &
                    0.029 & 
                    0.027 & 0.058 \\
        \quad AdaBoost & 0.537 & 0.539 & 
                         0.538 & 0.561 & 
                         0.011 & 
                         0.038 & 0.061 \\
        \quad Gradient boosting & 0.541 & 0.550 & 
                                  0.526 & 0.557 & 
                                  0.022 & 
                                  0.026 & 0.057 \\
        \midrule
				\multicolumn{7}{l}{\textsc{Deep learning}} \\
        \quad RNN & 0.583 & 0.548 & 
                    0.534 & 0.536 & 
                    0.020 & 
                    0.034 & 0.036 \\
        \quad LSTM & \bfseries 0.597 & \bfseries 0.573 & 
                     \bfseries 0.556 & 0.558 & 
                     \bfseries 0.045 & 
                     \bfseries 0.056 & 0.058 \\
        \quad LSTM with word embeddings & \bfseries 0.593 & \bfseries 0.579 & 
                                          \bfseries 0.566 & 0.551 &
                                          \bfseries 0.051 & 
                                          \bfseries 0.066 & 0.051 \\
				\midrule
				\multicolumn{7}{l}{\textsc{Transfer learning}} \\
        \quad RNN with pre-training & \bfseries 0.594 & 0.552 & 
                                      0.538 & 0.538 &
                                      0.024 & 
                                      0.038 & 0.038 \\
				\quad LSTM with pre-training & \bfseries 0.601 & \bfseries 0.576 & 
                                       \bfseries 0.563 & 0.552 & 
				                               \bfseries 0.048 & 
                                       \bfseries 0.063 & 0.052 \\
				\quad LSTM with pre-training & \bfseries 0.578 & \bfseries 0.578 & 
                                       \bfseries 0.583 & \bfseries 0.568 &
				                               \bfseries 0.050 & 
                                       \bfseries 0.083 & \bfseries 0.068 \\
				\quad and word embeddings \\
        \bottomrule
    \end{tabular}
\caption{Out-of-sample results from classifying the direction of the abnormal return. Values in bold indicate models that outperform both the na{\"i}ve baseline and traditional machine learning.}
\label{tbl:classification_abnormal_return}
\end{table}

\subsection{Regression: nominal returns}

While the previous section studied the accuracy in terms of classifying the direction of stock price changes, we now incorporate the nominal magnitude of the price adjustment. The corresponding results from the regression task are given in \Cref{tbl:regression_nominal_return}. With regard to the baseline models, only support vector regression yields favorable results in comparison to the na{\"i}ve approach (which represents the mean return of the training set). Its performance stems from choosing a radial basis function kernel and setting the cost to \num{0.05}.
While the previous section studied the accuracy in terms of classifying the direction of stock price changes, we now incorporate the nominal magnitude of the price adjustment. The corresponding results from the regression task are given in \Cref{tbl:regression_nominal_return}. With regard to the baseline models, only support vector regression yields favorable results in comparison to the na{\"i}ve approach. Its performance stems from choosing a radial basis function kernel and setting the cost to \num{0.05}.

The performance of the RNN is consistently inferior to both the na{\"i}ve approach and traditional machine learning. However, the LSTM outperforms the baselines on all metrics. It reduces the mean squared error of the random guess by \num{1.950} or \SI{5.08}{\percent}. Here word embeddings diminish the predictive performance, since they increase the mean squared error of the LSTM by \num{0.105}. 

Again, favorable results originate from transfer learning across all deep learning models, highlighting the benefits of the additional pre-training. Overall, the LSTM with pre-training and word embeddings performs best, decreasing the mean squared error by \num{2.053} (\ie \SI{5.34}{\percent}) compared to the na{\"i}ve approach.

\begin{table}
\hspace{-2.0cm}
\footnotesize
    \begin{tabular}{l SSSS SSS}
        \toprule
    	{\textbf{Method}} & 
		{\textbf{Training set}} & 
		\multicolumn{3}{c}{\textbf{Test set}} &
		\multicolumn{3}{c}{\textbf{\mcellt{Absolute error reduction\\ on test set over baseline}}}\\
		\cmidrule(l){2-2} \cmidrule(l){3-5} \cmidrule(l){6-8}
		& {\textbf{RMSE}} & 
		{\textbf{RMSE}} &
        {\textbf{MSE}} &
        {\textbf{MAE}} & 
		    {\textbf{RMSE}} &
        {\textbf{MSE}} &
        {\textbf{MAE}} \\ 
        \midrule
				\multicolumn{5}{l}{\textsc{Na{\"i}ve baseline}} \\
        \quad Mean return & 7.060 & 6.197 & 38.402 & 3.069 
                                   & \text{--} & \text{--} & \text{--} \\   
        \midrule
        \multicolumn{5}{l}{\textsc{Traditional machine learning}} \\
        \quad Ridge regression & 6.765 & 6.127 & 37.541 & 3.114 
                                      & -0.070 & -0.861 & 0.045 \\
        \quad Lasso & 6.918 & 6.122 & 37.486 & 3.089 
                            & -0.075 & -0.916 & 0.020 \\ 
        \quad Elastic net & 6.892 & 6.108 & 37.308 & 3.091 
                                  & -0.089 & -1.094 & 0.022 \\        
        \quad Random forest  & 6.873 & 6.145 & 37.761 & 3.111 
                                     & -0.052 & -0.641 & 0.042 \\
        \quad SVR  & 6.890 & 6.171 & 38.081 & 3.058 
                           & -0.026 & -0.321 & -0.011 \\
        \quad AdaBoost & 7.994 & 7.282 & 53.028 & 4.837 
                               & 1.085 & 14.626 & 1.768 \\
        \quad Gradient boosting & 6.872 & 6.146 & 37.773 & 3.111 
                                        & -0.051 & -0.629 & 0.042 \\
        \midrule
				\multicolumn{5}{l}{\textsc{Deep learning}} \\
        \quad RNN & 6.859 & 6.139 & 37.685 & 3.102 
                          & -0.058 & -0.717 & 0.033 \\
        \quad LSTM & 6.892 & \bfseries 6.038 & \bfseries 36.452 & \bfseries 3.024 
                           & \bfseries -0.159 & \bfseries -1.950 & \bfseries -0.045 \\
        \quad LSTM with word embeddings & 6.954 & \bfseries 6.046 & \bfseries 36.557 & \bfseries 3.043 
                                                & \bfseries -0.151 & \bfseries -1.845 & \bfseries -0.026 \\
				\midrule
				\multicolumn{5}{l}{\textsc{Transfer learning}} \\
        \quad RNN with pre-training & 6.875 & \bfseries 6.101 & \bfseries 37.222 & 3.099 
                                            & \bfseries -0.096 & \bfseries -1.180 & 0.030 \\
				\quad LSTM with pre-training & 6.887 & \bfseries 6.036 & \bfseries 36.439 & \bfseries 3.020 
				                                     & \bfseries -0.161 & \bfseries -1.963 & \bfseries -0.049 \\
				\quad LSTM with pre-training & 6.876 & \bfseries 6.029 & \bfseries 36.349 & \bfseries 3.011 
				                                     & \bfseries -0.168 & \bfseries -2.053 & \bfseries -0.058 \\
				\quad and word embeddings \\
        \bottomrule
    \end{tabular}
\caption{Out-of-sample results from regressing the nominal return. Bold values indicate models that outperform all baselines (\ie na{\"i}ve and traditional machine learning).}
\label{tbl:regression_nominal_return}
\end{table}

\subsection{Regression: abnormal returns}

\Cref{tbl:regression_abnormal_return} evaluates the regression task with abnormal returns, which corrects for confounding market movements. Here the elastic net achieves the lowest mean squared error among the traditional machine learning models, amounting to \num{37.614}, which is \num{-0.930} below the na{\"i}ve benchmark. 

Among deep learning approaches, the LSTM with word embeddings achieves the lowest mean squared error, outperforming the na{\"i}ve approach by an absolute reduction of \num{2.281}. This model corresponds to choosing a learning rate of 0.0005 and \num{50}-dimensional vectors. In contrast to the previous regression task with nominal returns, we observe mixed results when incorporating word embeddings: doing so decreases the mean squared error, but increases the mean absolute error. 

Using transfer learning further improves the prediction performance of LSTMs. It shrinks the mean squared error by an additional \num{0.110} and \num{0.083} for the classical LSTM model and the LSTM model using word embeddings, respectively. Overall, the LSTM with pre-training and word embeddings outperforms the na{\"i}ve approach by an absolute value of \num{2.364} (\ie \SI{6.1}{\percent}) in terms of mean squared error.

\begin{table}
\hspace{-2.0cm}
\footnotesize
    \begin{tabular}{l SSSS SSS}
        \toprule
    	{\textbf{Method}} & 
		{\textbf{Training set}} & 
		\multicolumn{3}{c}{\textbf{Test set}} &
		\multicolumn{3}{c}{\textbf{\mcellt{Absolute error reduction\\ on test set over baseline}}} \\
		\cmidrule(l){2-2} \cmidrule(l){3-5} \cmidrule(l){6-8}
		& {\textbf{RMSE}} & 
		    {\textbf{RMSE}} &
        {\textbf{MSE}} &
        {\textbf{MAE}} &
        {\textbf{RMSE}} &
        {\textbf{MSE}} &
        {\textbf{MAE}} \\ 
        \midrule
				\multicolumn{5}{l}{\textsc{Na{\"i}ve baseline}} \\
        \quad Mean abnormal return & 7.102 & 6.208 & 38.544 & 3.126
                                   & \text{--} & \text{--} & \text{--} \\   
        \midrule
        \multicolumn{5}{l}{\textsc{Traditional machine learning}} \\
        \quad Ridge regression & 6.908 & 6.153 & 37.860 & 3.168 
                                       & -0.055 & -0.684 & 0.042 \\
        \quad Lasso & 6.992 & 6.153 & 37.860 & 3.166 
                            & -0.055 & -0.684 & 0.040 \\ 
        \quad Elastic net & 6.956 & 6.133 & 37.614 & 3.144 
                                  & -0.075 & -0.930 & 0.018 \\        
        \quad Random forest  & 6.927 & 6.173 & 38.106 & 3.176 
                                     & -0.035 & -0.438 & 0.050\\
        \quad SVR  & 6.906 & 6.183 & 38.235 & 3.122 
                           & -0.025 & -0.309 & -0.004 \\
        \quad AdaBoost & 8.046 & 7.159 & 51.251 & 4.704 
                               & 0.951 & 12.707 & 1.578 \\
        \quad Gradient boosting & 6.923 & 6.173 & 38.106 & 3.176 
                                        & -0.035 & -0.438 & 0.050 \\
        \midrule
				\multicolumn{5}{l}{\textsc{Deep learning}} \\
        \quad RNN & 6.966 & 6.169 & 38.062 & 3.162 
                          & -0.039 & -0.482 & 0.036 \\
        \quad LSTM & \bfseries 6.873 & \bfseries 6.030 & \bfseries 36.364 & \bfseries 3.118 
                                     & \bfseries -0.178 & \bfseries -2.180 & \bfseries -0.008\\
        \quad LSTM with word embeddings & \bfseries 6.904 & \bfseries 6.022 & \bfseries 36.263 & \bfseries 3.127 
                                                          & \bfseries -0.186 & \bfseries -2.281 & 0.001 \\
				\midrule
				\multicolumn{5}{l}{\textsc{Transfer learning}} \\
        \quad RNN with pre-training & 6.934 & 6.133 & 37.614 & 3.158 
                                            & -0.064 & -0.788 & 0.089 \\
				\quad LSTM with pre-training & \bfseries 6.845 & \bfseries 6.021 & \bfseries 36.254 & \bfseries 3.109 
				                                               & \bfseries -0.187 & \bfseries -2.290 & \bfseries -0.017 \\
				\quad LSTM with pre-training & \bfseries 6.687 & \bfseries 6.015 & \bfseries 36.180 & \bfseries 3.104 
				                                               & \bfseries -0.193 & \bfseries -2.364 & \bfseries -0.022 \\
				\quad and word embeddings \\
        \bottomrule
    \end{tabular}
\caption{Out-of-sample results from regressing the abnormal return. Values in bold indicate an improvement compared to all baselines (\ie na{\"i}ve and traditional machine learning).}
\label{tbl:regression_abnormal_return}
\end{table} 

\subsection{Sensitivity analysis}

We now investigate the sensitivity of our models to modifications in their parameters (see online appendix for details). We first explore the effect of introducing $n$-grams within traditional machine learning models and varying the size of $n$. In short, in the classification task with abnormal returns, bag-of-words models indicate mixed results when changing the length of $n$-grams (\ie bigrams and trigrams). For instance, the test accuracy of ridge regression improves by \num{0.3} percentage points when utilizing bigrams compared to unigrams, but decreases by \num{0.4} percentage points when applying trigrams. On the contrary, the test accuracy of AdaBoost decreases by \num{0.3} percentage points for bigrams, but increases by \num{0.3} percentage points for trigrams. Altogether, we observe no clear pattern that can guide our choice of $n$ and, more importantly, none of our experiments resulted in a performance that is superior to LSTMs.

We empirically investigate whether the deep neural network can store long sequences that span a complete sentence or even more extensive text passages. For this purpose, we shorten each document and merely extract the first words. We then train a deep neural network with this shortened text fragment in order to determine whether the deep neural network with the complete documents yields a superior predictive performance. Due to space constraints, we only detail the regression task with abnormal returns for an LSTM with word embeddings. As a result, the RMSE on the test set attains a value of \num{6.162} when considering merely the first sentence, \num{6.184} when restricting the document to the first 50 words and \num{6.114} for the first 100 words. Here we observe that all experiments result in a worse predictive performance than the network with the complete documents (RMSE of \num{6.022}). This implies that the neural network can learn to store even long sequences in its weights.

Deep learning usually works as a black-box approach and, as a remedy, we contribute to explanatory insights as follows: we draw upon the finance-specific dictionary from Loughran-McDonald that comprises terms labeled as either positive or negative, where the underlying categorization stems from subjective human ratings. We then treat each word as a single document and insert them as input into our deep neural network. The resulting predictions allow us to infer whether a word links to a positive or negative market reaction. In other words, the prediction scores the polarity of the words and specifies how markets perceive them. We show an excerpt in \Cref{tbl:word_predictions}, while the supplementary materials provide the complete list.

\begin{table}
\footnotesize
\centering
\begin{tabular}{llS}
\toprule
{\textbf{Entry}} &
{\textbf{Label}} & 
{\textbf{Predicted score}} \\ 
\midrule
absence & Negative & -0.176 \\
abuse & Negative & -0.034 \\
achieve & Positive & 0.338 \\
adequately & Positive & 0.284 \\
advantage & Positive & 0.256 \\
\ldots & \ldots & \ldots \\
\bottomrule
\end{tabular}
\caption{Standardized predictions for individual terms from the Loughran-McDonald finance-specific word list. The results stem from an LSTM with word embeddings for the regression task with abnormal returns. The complete list is reported in the supplements.}
\label{tbl:word_predictions}
\end{table}

\section{Discussion}
\label{sec:discussion}

\subsection{Comparison}

We additionally compare our predictive performance of stock market movements to earlier publications studying the same financial disclosures. However, the results are often not comparable, as different papers utilize additional (subjective) filter rules, report accuracies instead of balanced accuracies, incorporate other splits into training and test sets, or neglect to perform time-series cross-validation. We refer to previous literature overviews~\cite{Hagenau.2013,Nassirtoussi.2014} for a comparison of predictive accuracy across different news sources.


In short, the SVM-based approach in \cite{Hagenau.2013} is fed with a different time frame (starting in 1997), which results in a skewer distribution of positive and negative labels, with \SI{58.3}{\percent} of them being positive. Since the authors do not report a balanced accuracy, we cannot make a fair comparison. Furthermore, they train their classifiers without temporal distinction, resulting in an over estimated performance. Hence, we replicated their experiments with 2-gram and 3-gram SVMs using our dataset, training processes and evaluation metrics. However, the performance of the SVM-based approach is substantially inferior to that of deep learning. The recursive autoencoder in \cite{Feuerriegel.2016} splits the announcements into three classes (up, down or steady) according to the abnormal return and then discards the steady samples a priori. Moreover, their approach relies merely upon headlines and reports the accuracy (\SI{56}{\percent}) instead of the balanced accuracy. Furthermore, their accuracy is lower than ours due to the advanced deep learning architectures. By additionally incorporating the discourse structure, the authors of \cite{Markle.2017} achieve a balanced accuracy amounting to \SI{54.32}{\percent} for the same dataset. However, this performance is yet again inferior to that of deep and transfer learning.


\subsection{Generalizability and limitations}

The aforementioned models based on deep learning are not limited to sentiment analysis or natural language processing, but can be beneficial in any task of advanced complexity, such as time series prediction, voice control or information retrieval. In this respect, deep learning can help to encode context information that spans multiple words or even sentences.

The majority of deep learning architectures are trained in a supervised fashion and thus need a sufficiently large labeled dataset. This requirement can be partially relaxed by transfer learning, which first performs representation learning on a different, but related, dataset and then solves problems regarding the actual data. For instance, decision support from financial disclosures can benefit from systems trained on a different type of news. To do so, one first tunes the parameters on the basis of general finance-related narratives in order to acquire a basic understanding of language in a finance-specific context and, in a next step, tailors the weights in the output layer to the particular problem. 

In comparison to classical machine learning tasks, accurate predictions based on financial news are still difficult to obtain due to the complexity of natural language and the efficiency of markets where historic prices contribute only marginally to explaining future returns~\cite{Tetlock.2007}. Both difficulties underline the necessity for more complex models in the field of deep learning. Nevertheless, even minor improvements in the predictive performance can result in a considerable economic impact. For this reason, we assume a portfolio of \$\num{1000}, one year with 200 trading days and each with a single disclosure that triggers a log-return of \SI{5}{\percent}. By utilizing a random guess in the prediction engine, this would result in an expected final portfolio of \$\num{1000}. A 51\,\% accuracy increases the monetary value of the portfolio considerably, since the portfolio now attains a log-return of \SI{10}{\percent} over the course of the year. 

\subsection{Implications for management}

Deep learning is applicable to the improvement of decision support in many core areas of organizations and businesses, such as recommender systems, question-answering mechanisms and customer support. To further augment potential use cases, the long short-term memory model enables the processing of sequential data, often with unprecedented performance. This model thus allows one to bolster existing tool chains, where traditional predictive models will soon be replaced by deep architectures.

This substitution, however, can be challenging in practical application due to rapid advances in the underlying software libraries. As an example, the above results required considerable adjustments in TensorFlow and Theano, such as regularization to avoid overfitting of the networks. While many pre-trained networks are available for image-related tasks, this is often not the case for natural language processing, which is why we published our trained networks as open-source.


\subsection{Implications for research}

This work demonstrates the achievements of deep learning in relation to decision support systems and, at the same time, presents opportunities for research aimed at the enhancement of transfer learning in natural language processing. Transfer learning has been predominantly applied to image-related tasks; however, empirical results are scarce when it comes to natural language processing. Common obstacles derive from the fact that large, pre-assembled datasets are often not readily available. The incorporation of these large-scale corpora, however, is essential to building powerful models.
 Therefore, future research might adapt the idea behind the ImageSet dataset and publish extremely large unlabeled and labeled datasets for text classifications. 

\section{Conclusion and outlook}
\label{sec:conclusion}

Financial disclosures greatly aid investors and automated traders in deciding whether to exercise ownership in stocks. While humans are usually able to interpret textual content correctly, computerized decision support systems struggle with the complexity and ambiguity of natural language. 

This paper analyzes the switch from traditional bag-of-words models to deep, non-linear neural networks. Each of the neural networks comprises more than \num{500000} parameters that help in making accurate predictions. Thereby, we contribute to the existing literature by showing how deep learning can enhance financial decision support by explicitly incorporating word order, context-related information and semantics. For this purpose, we engage in the task of predicting stock market movements subsequent to the disclosure of financial materials. Our results show that long short-term memory models can outperform all traditional machine learning models based on the bag-of-words approach, especially when we further pre-train word embeddings with transfer learning. We thus identify two critical ingredients for superior predictive performance, namely being able to infer context-dependent information from ordered sequences of words and capturing highly non-linear relationships. Yet the configuration of deep neural networks represents a challenging task, as it still requires extensive parameter tuning to achieve favorable results. With regard to news-based predictions, it is an interesting question for future research to further detail the gains in predictive performance from deep learning for intraday data (including potential latency effects) and in the long run. 

We expect that deep learning will soon expand beyond the realm of academic research and the rather limited number of firms that specialize in predictive analytics, especially as decision support systems can benefit from deep learning in multiple ways. First of all, deep learning can learn to incorporate context information from sequential data. Second, competition will drive firms and organizations towards using more powerful architectures in predictive tasks and, in this regard, deep neural networks with transfer learning often represent the status quo.


\vspace*{-0.4cm}
\section*{Acknowledgments}

{\footnotesize 
 We thank Ryan Grabowski for his proof-reading.
}


\section*{References}
\vspace*{-0.4cm}

\bibliographystyle{model1-num-names}
\bibliography{literature}







\end{document}


\pagenumbering{gobble}%
\appendix

\section{Tuning parameters}

\Cref{tbl:tuning_parameter} reports the tuning parameters used in our grid search.

\begin{table}
\hspace{-3.8cm}
\scriptsize
\begin{tabular}{lll rrrr}
\toprule
\textbf{Predictive model} & \textbf{Tuning parameters} & \textbf{Tuning range} & \multicolumn{4}{c}{\textbf{Final parameter choice}} \\ 
\cline{4-7}
& & & \textbf{C/R} & \textbf{C/AR} & \textbf{R/R} & \textbf{R/AR} \\
\midrule
Ridge regression & Regularization strength $\alpha$ & 
1, 0.99, 0.95, 0.9, \ldots, 0.1, 0.05, 0.01 & 0.9 & 0.99 & 0.9 & 0.85 \\
& Number $n_{\max}$ of selected features & 200, 300, 400, 500, 1000, all & 300 & 400 & 500 & 500 \\
& Stemming & enabled, disabled & enabled & disabled & enabled & enabled \\[0.5em]
Lasso & Regularization strength $\alpha$ & 
1, 0.99, 0.95, 0.9, \ldots, 0.1, 0.05, 0.01 & --- & --- & 0.9 & 0.8 \\
& Number $n_{\max}$ of selected features & 200, 300, 400, 500, 1000, all & --- & --- & all & 1000 \\
& Stemming & enabled, disabled & --- & --- & enabled & enabled \\[0.5em]
Elastic net & Ratio of $L_1$- and $L_2$-penalties $\alpha_1$, $\alpha_2$ &
1, 0.99, 0.95, 0.9, \ldots, 0.1, 0.05, 0.01 & --- & --- & 0.95 & 0.95 \\
& Number $n_{\max}$ of selected features & 200, 300, 400, 500, 1000, all & --- & --- & 500 & 500 \\
& Stemming & enabled, disabled & --- & --- & enabled & enabled \\[0.5em]
Random forest & Number of randomly-sampled variables & 1, 2, 3, 5, 7 & 3 & 5 & 3 & 2 \\ 
& Number of trees & 100, 200, 500, 1000 & 1000 & 500 & 200 & 500 \\ 
& Maximum depth of trees & 1, 5, 10, 20, 50, 100, None & None & 20 & None & 100 \\
& Number $n_{\max}$ of selected features & 200, 300, 400, 500, 1000, all & 400 & all & 300 & 300 \\
& Stemming & enabled, disabled & disabled & disabled & enabled & enabled \\[0.5em]
AdaBoost & Maximum number of trees & 1, 5, 10, 20, 50, 100 & 50 & 100 & 50 & 20 \\ 
& Learning rate & 1, 0.99, 0.95, 0.9, \ldots, 0.1, 0.05, 0.01 & 0.15 & 0.2 & 0.05 & 0.05 \\
& Number $n_{\max}$ of selected features & 200, 300, 400, 500, 1000$^\dagger$ & 1000 & all & 300 & 300 \\
& Stemming & enabled, disabled & enabled & enabled & enabled & enabled \\[0.5em]
Gradient boosting & Maximum number of weak learners & 100, 200, 500, 1000 & 500 & 200 & 100 & 200 \\
& Loss function & [least squares, least absolute, huber] & least squares & least squares & least squares & least squares \\ 
& Learning rate & 1, 0.99, 0.95, 0.9, \ldots, 0.1, 0.05, 0.01 & 0.1 & 0.05 & 0.1 & 0.1 \\
& Number $n_{\max}$ of selected features & 200, 300, 400, 500$^\dagger$ & 300 & 400 & 200 & 300 \\
& Stemming & enabled, disabled & enabled & enabled & enabled & enabled \\[0.5em]
SVM & Kernel function & Linear, radial, poly & linear & linear & radial & radial \\
& Cost $C$ & 1.0, 0.99, 0.95, 0.9, \ldots, 0.1, 0.05, 0.01 & 0.15 & 0.15 & 0.05 & 0.1 \\
& Number $n_{\max}$ of selected features & 200, 300, 400, 500, 1000, all & 300 & 200 & 200 & 500 \\
& Stemming & enabled, disabled & enabled & disabled & enabled & disabled \\[0.5em]
RNN & Word embedding size & 30, 40, \ldots, 100 & 40 & 30 & 50 & 40 \\
& Learning rate & 0.0001, 0.0005, \ldots, 0.0095, 0.01 & 0.0005 & 0.0005 & 0.005 & 0.001 \\
& Stemming & enabled, disabled & disabled & disabled & disabled & disabled \\[0.5em]
LSTM & Word embedding size & 30, 40, \ldots, 100 & 50 & 40 & 60 & 60 \\
& Learning rate & 0.0001, 0.0005, \ldots, 0.0095, 0.01 & 0.0005 & 0.005 & 0.001 & 0.001 \\
& Stemming & enabled, disabled & disabled & disabled & disabled & disabled \\[0.5em]
LSTM with & Word embedding size & 30, 40, \ldots, 100 & 60 & 70 & 70 & 50 \\
word embeddings & Learning rate & 0.0001, 0.0005, \ldots, 0.0095, 0.01 & 0.01 & 0.005 & 0.005 & 0.005 \\
& Stemming & enabled, disabled & disabled & disabled & disabled & disabled \\ 
\bottomrule
\multicolumn{3}{l}{$^\dagger$ Skipped larger values due to high computational time} 
\end{tabular}
\caption{Parameters tuned via time-series cross-validation in order to find the best model. Here we refer to the different experiments as follows: classification with nominal returns (C/R) and abnormal returns (C/AR); regression with nominal (R/R) and abnormal returns (R/AR).}
\label{tbl:tuning_parameter}
\end{table} 

\newpage
\section{Sensitivity analysis: $n$-grams}

In the following, \Cref{tbl:bigrams} shows the effect of utilizing bigrams within traditional machine learning models (with stemming only) to classify the direction of abnormal return. As a result, ridge regression performs best across all approaches and metrics, outperforming the na{\"i}ve approach by \num{4.8} percentage points in terms of balanced accuracy. Among non-linear models, random forest performs best with a balanced accuracy of \SI{54.1}{\percent}, amounting to an increased performance of \num{4.1} percentage points. Overall, we observe no clear evidence that either unigrams or bigrams performs better.

\begin{table}
\centering
\footnotesize
    \begin{tabular}{l SSSS}
		\toprule
		{\textbf{Method}} & 
		{\textbf{Training set}} & 
		\multicolumn{3}{c}{\textbf{Test set}} \\
		\cmidrule(l){2-2} \cmidrule(l){3-5}
		& {\textbf{Accuracy}} &
		{\textbf{Accuracy}} & 
		{\textbf{\mcellt{Balanced\\ accuracy}}} &
		{\textbf{AUC}} \\
        \midrule
        \quad Na{\"i}ve baseline (majority class)  & 0.542 & 0.528 & 
                              0.500 & 0.500 \\   
        \midrule
        \quad Ridge regression & 0.540 & \bfseries 0.552 & 
                              \bfseries 0.548 & \bfseries 0.560 \\
        \quad Lasso & 0.542 & 0.528 & 
                              0.500 & 0.500 \\ 
        \quad Elastic net & 0.542 & 0.528 & 
                              0.500 & 0.500 \\       
        \quad Random Forest & \bfseries 0.554 & 0.548 & 
                              0.541 & 0.555 \\
        \quad SVM & 0.548 & 0.536 & 
                              0.523 & 0.549 \\
        \quad AdaBoost & 0.547 & 0.535 & 
                              0.525 & 0.551 \\
        \quad Gradient Boosting & 0.552 & 0.535 & 
                              0.514 & 0.534 \\
       \bottomrule
    \end{tabular}
\caption{Out-of-sample results from baselines utilizing bigrams to classify the direction of abnormal return.}
\label{tbl:bigrams}
\end{table}

\Cref{tbl:trigrams} reports the results of traditional models with stemming only when incorporating trigrams. Again, ridge regression and random forest perform well, beating the na{\"i}ve benchmark by \num{4.1} and \num{4.0} percentage points, respectively. Again, there is no direct recommendation on whether to prefer unigrams or trigrams

\begin{table}
\hspace{1.4cm}
\footnotesize
    \begin{tabular}{l SSSS}
		\toprule
		{\textbf{Method}} & 
		{\textbf{Training set}} & 
		\multicolumn{3}{c}{\textbf{Test set}} \\
		\cmidrule(l){2-2} \cmidrule(l){3-5}
		& {\textbf{Accuracy}} &
		{\textbf{Accuracy}} & 
		{\textbf{\mcellt{Balanced\\ accuracy}}} &
		{\textbf{AUC}} \\
        \midrule
        \quad Na{\"i}ve baseline (majority class) & 0.542 & 0.528 & 
                              0.500 & 0.500 \\   
        \midrule
        \quad Ridge regression & 0.538 & 0.545 & 
                              \bfseries 0.541 & \bfseries 0.565 \\
        \quad Lasso & 0.542 & 0.528 & 
                              0.500 & 0.500 \\ 
        \quad Elastic net & 0.542 & 0.528 & 
                            0.500 & 0.500 \\       
        \quad Random Forest & \bfseries 0.554 & \bfseries 0.547 & 
                              0.540 & 0.552 \\
        \quad SVM & 0.547 & 0.540 & 
                    0.527 & 0.549 \\
        \quad AdaBoost & 0.544 & 0.541 & 
                         0.532 & 0.552 \\
        \quad Gradient Boosting & 0.552 & 0.537 & 
                                  0.516 & 0.536 \\
       \bottomrule
    \end{tabular}
\caption{Out-of-sample results from baselines utilizing trigrams to classify the direction of abnormal return.}
\label{tbl:trigrams}
\end{table}

\newpage
\section{Explanatory insights for finance-specific words}

\Cref{tbl:McDonald_pos} and \Cref{tbl:McDonald_neg} report the standardized results of the LSTM for classifying abnormal returns with pre-training and word embeddings when predicting sentiment scores of single words from the Loughran-McDonald dictionary~\cite{Loughran.2011}. We observe several differences across inflected word forms. This is partially owed to the fact that these words are not part of Glove and thus become initialized randomly. We thus repeat the analysis with an LSTM that is trained on word stems; see \Cref{tbl:McDonald_pos_stemming} and \Cref{tbl:McDonald_neg_stemming}.

\subsection{Inflected words}

\begin{center}
\footnotesize
\begin{longtable}{llS[round-mode=places,round-precision=3]}
\toprule
{\textbf{Entry}} &
{\textbf{Label}} & 
{\textbf{Predicted score}} \\ 
\midrule
achieve  & Positive & 0.338 \\
achieved  & Positive & 0.478 \\
achievement  & Positive & 0.282 \\
achievements  & Positive & 0.216 \\
achieves  & Positive & 0.08 \\
achieving  & Positive & 0.054 \\
adequately  & Positive & 0.284 \\
\end{longtable}
\end{center}

\begin{center}
\footnotesize
\begin{longtable}{llS[round-mode=places,round-precision=3]}
\toprule
{\textbf{Entry}} &
{\textbf{Label}} & 
{\textbf{Predicted score}} \\ 
\midrule
 absence  & Negative & -0.176 \\
 abuse  & Negative & -0.034 \\
 adverse  & Negative & 0.028 \\
 adversely  & Negative & -0.136 \\
 against  & Negative & -0.092 \\
 allegations  & Negative & -0.214 \\
 alleged  & Negative & -0.444 \\
\end{longtable}
\end{center}

\newpage
\subsection{Word stems}

\begin{center}
\footnotesize
\begin{longtable}{llS[round-mode=places,round-precision=3]}
\toprule
{\textbf{Stem}} &
{\textbf{Label}} & 
{\textbf{Predicted score}} \\ 
\midrule
achiev & Positive & 0.185 \\
adequ & Positive & 0.287 \\
advanc & Positive & 0.055 \\
advantag & Positive & 0.055 \\
allianc & Positive & 0.173 \\
assur & Positive & -0.069 \\
attain & Positive & 0.164 \\
\end{longtable}
\end{center}

\begin{center}
\footnotesize
\begin{longtable}{llS[round-mode=places,round-precision=3]}
\toprule
{\textbf{Stem}} &
{\textbf{Label}} & 
{\textbf{Predicted score}} \\ 
\midrule
absenc & Negative & -0.137 \\
abus & Negative & 0.022 \\
advers & Negative & 0.032 \\
against & Negative & -0.090 \\
alleg & Negative & -0.322 \\
antitrust & Negative & 0.627 \\
bad & Negative & 0.007 \\
\end{longtable}
\end{center}



\section*{References}

\bibliographystyle{model1-num-names}
\bibliography{literature}





